\documentclass[runningheads]{llncs}
\usepackage[T1]{fontenc}
\usepackage{graphicx}
\usepackage{subcaption}
\usepackage{booktabs}

\usepackage{hyperref}
\usepackage{amsmath}
\usepackage{caption}
\usepackage{changes}

\usepackage{xcolor}

\usepackage{tcolorbox}
\tcbuselibrary{breakable}
\newtcolorbox{definitionbox}{
  breakable,
  colback=gray!5,
  colframe=gray!40,
  boxrule=0.5pt,
  arc=2pt,
  left=4pt,
  right=4pt,
  top=4pt,
  bottom=4pt
}

\usepackage{tikz}
\definecolor{clusterred}{RGB}{254, 192, 193}
\definecolor{clusterorange}{RGB}{254, 217, 193}
\definecolor{clusteryellow}{RGB}{255, 232, 159}
\definecolor{clusterpurple}{RGB}{213, 193, 225}
\definecolor{clusterblue}{RGB}{192, 212, 230}
\definecolor{clustergreen}{RGB}{209, 244, 208}
\definecolor{colorbrown}{RGB}{165, 42, 42}       % brown
\definecolor{colorpink}{RGB}{255, 192, 203}      % pink
\definecolor{colorblue2}{RGB}{70, 130, 180}       % steelblue
\definecolor{colororange2}{RGB}{255, 165, 0}      % orange
\definecolor{colorgreen}{RGB}{0, 128, 128}        % teal

\newcommand{\colorcircle}[2]{\raisebox{-2pt}{\begin{tikzpicture}[scale=0.2] \node[circle, fill=#1, font=\footnotesize, inner sep=0.35mm]{#2} ; \end{tikzpicture}}}
\newcommand{\llod}{L^\text{LOD}}
\newcommand{\ltxt}{L^\text{TXT}}

\begin{document}
\title{Which Are the Low-Resource Languages of the Semantic Web?}
\titlerunning{Which Are the Low-Resource Languages of the Semantic Web?}
\author{Ndeye-Emilie Mbengue\inst{1}
\orcidID{0009-0002-1289-6417}, 
Pierre Monnin\inst{1}\orcidID{0000-0002-2017-8426},
Miguel Couceiro\inst{2}\orcidID{0000-0003-2316-7623}, 
Fabien Gandon\inst{1}\orcidID{0000-0003-0543-1232}}
\authorrunning{N.-E. Mbengue et al.}

\institute{Université Côte d'Azur, Inria, CNRS, I3S, Sophia-Antipolis, France \\
\email{\{ndeye-emilie.mbengue, pierre.monnin, fabien.gandon\}@inria.fr}
\and INESC-ID, Instituto Superior Técnico, Universidade de Lisboa, Lisbon, Portugal \\
\email{miguel.couceiro@inesc-id.pt}}
\maketitle   

%%%%%%%%%%%%%%%%%%%%%%%%%%%%%%%%%%%%%%%%%%%%%%%%%%%%%
\begin{abstract}Emerging digital technologies are exacerbating the existing divide in Open Access Data (OAD) between high- and low-resource languages, excluding many communities from the global digital transformation.
Multilingual Linked Open Data Knowledge Graphs (LOD KGs) could contribute to mitigating this divide through cross-lingual transfer; however, no clear quantitative definition of low-resource languages has yet been established in the context of LOD KGs. 
In this poster, we present a methodology to analyze the distribution of languages across LOD KGs and propose a preliminary multi-level categorization based on DBpedia, BabelNet, and Wikidata. This categorization is leveraged to bring a formal definition of low-, high-, and medium-resource languages that could be later leveraged to select cross-lingual transfer candidates.
\end{abstract}

\section{Introduction}

The adaptation of digital tools to the diverse linguistic and cultural specificities is essential to ensure equitable digital access. 
However, the existing gap between high- and low-resource languages in open data~\cite{Viksna2022AssessingMO}, upon which many emerging technologies such as Generative AI are built, contributes to digital access inequalities.
For instance, the lack of corpora in a given language prevents the development of language models for that language, thus limiting some tasks such as text generation or question-answering possibilities~\cite {helm_diversity_2023}.
%As a result, numerous languages and cultures remain critically underserved in essential sectors such as healthcare and education~\cite{helm_diversity_2023}.
%One approach to mitigate this divide lies in leveraging multilingual Linked Open Data Knowledge Graphs (LOD KGs) through cross-lingual transfer and logical reasoning. 
To address this gap, one could rely on multilingual Linked Open Data Knowledge Graphs (LOD KGs) that provide an abstract and unified representation of multilingual and heterogeneous data, which facilitates cross-lingual transfer. Improving LOD KGs language coverage can thus help in: $(i)$ reducing the gap in OAD by generating new data in low-resource languages, $(ii)$ providing better digital access and fairness, since they also underpin many digital applications, including search engines, recommendation systems, or GraphRAG information systems.
%Knowledge graphs are graphs of data intended to accumulate and convey structured knowledge where nodes represent entities and edges represent relationships between them~\cite{2021Hogan}, that can be leveraged with logical reasoning and a transfer mechanism. Consequently, multilingual LOD KGs offer the opportunity to enhance both the quality and quantity of available low-resource language data.
Improving LOD KGs can rely on cross-lingual transfer, which requires first identifying suitable low-resource language targets, along with corresponding high-resource languages to transfer from. Nevertheless, there is neither a clear definition of low-resource languages in the context of LOD KGs, nor a quantitative framework for categorizing languages in LOD, motivating the following research questions: \textbf{How to categorize languages according to their digital coverage in LOD?} \textbf{How to define low-resource languages in LOD?} 
%And to extend the traditional binary classification between low and high resource languages: How to develop a fine-grained categorization of languages based on their level of digital coverage in Linked Open Data?
%To explore this question, we (i) present a brief literature review on existing approaches used to characterize language digital coverage, (ii) propose a methodology for quantifying language resources from a Linked Open Data perspective, and (iii) apply this methodology to three multilingual knowledge graphs, providing analysis on the language distribution landscape. This analysis is supported by a Web tool accessible at \url{https://nembengue.github.io/language_digital_coverage_lod/}.

\section{How to Categorize Language Digital Coverage in LOD?}

\begin{figure}
    \centering
   \begin{subfigure}[b]{0.41\textwidth}
        \centering
        \includegraphics[width=0.85\linewidth]{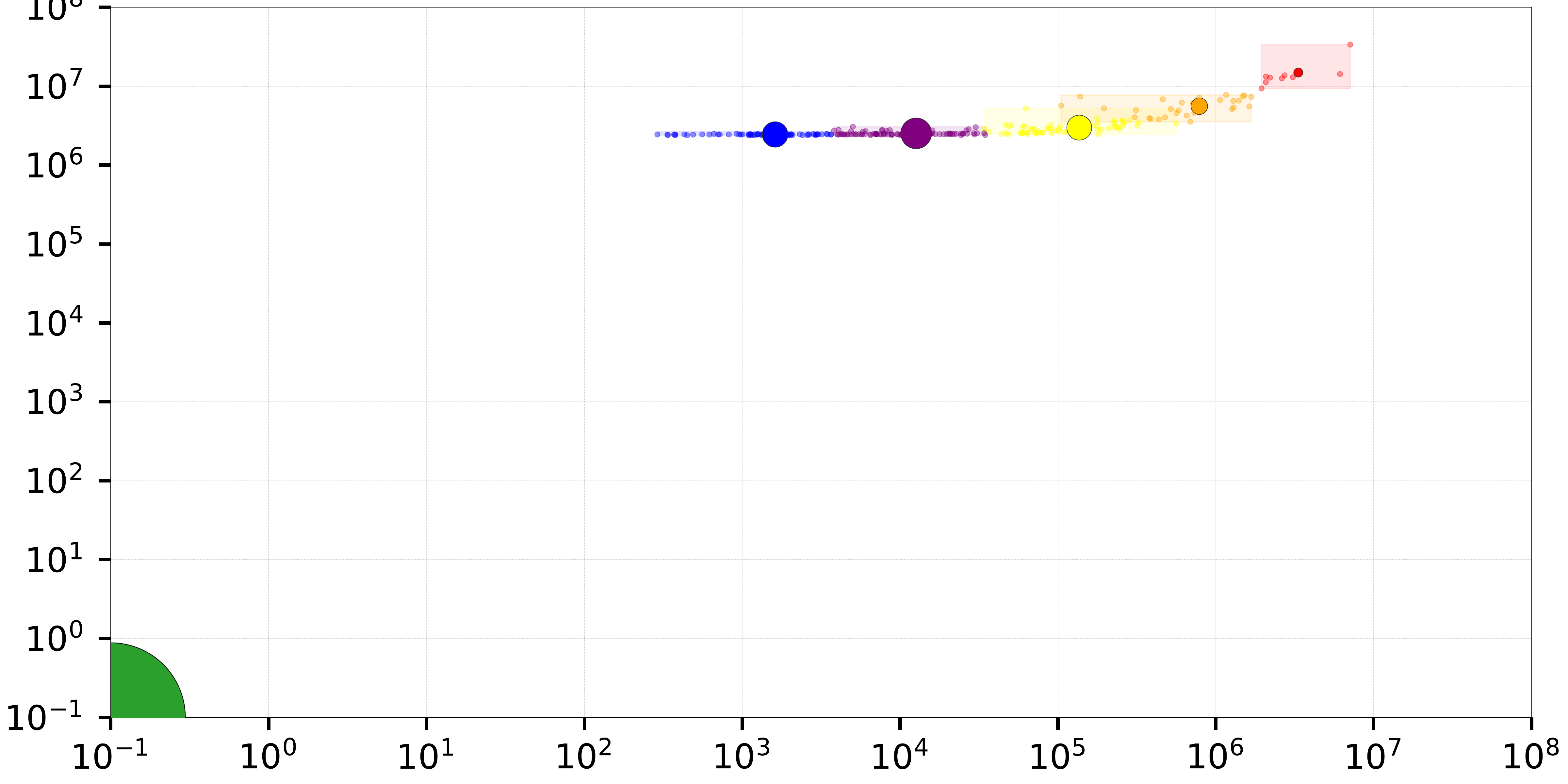}
        \caption{BabelNet-based categories.}
        \label{fig:babelnet}
    \end{subfigure}
    \qquad
    \begin{subfigure}[b]{0.41\textwidth}
        \centering 
        \includegraphics[width=0.85\linewidth]{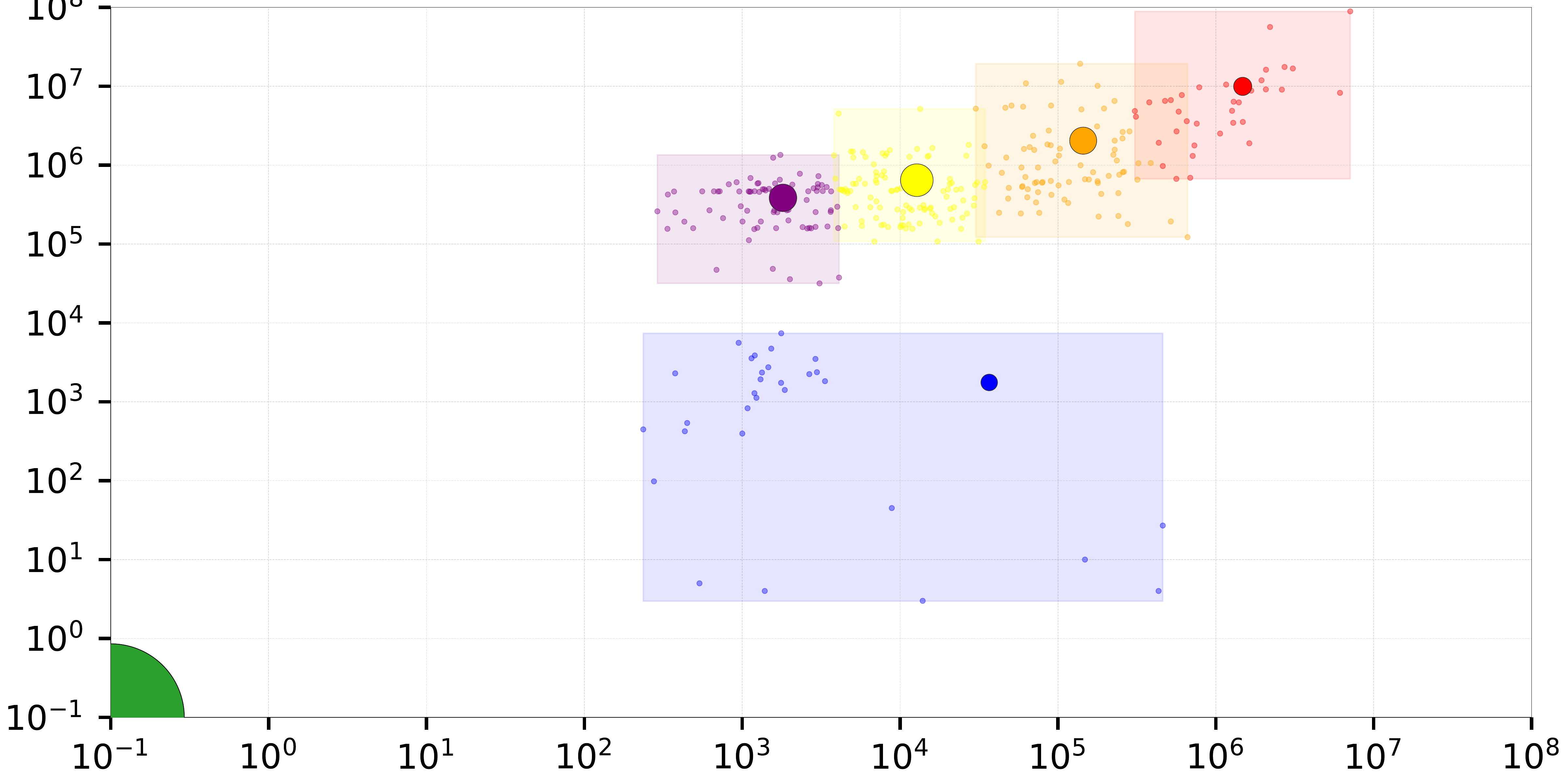}
        \caption{Wikidata-based categories.}
        \label{fig:wikidata}
    \end{subfigure}
    % Row 2
    \begin{subfigure}[b]{0.41\textwidth}
        \centering 
        \includegraphics[width=0.85\linewidth]{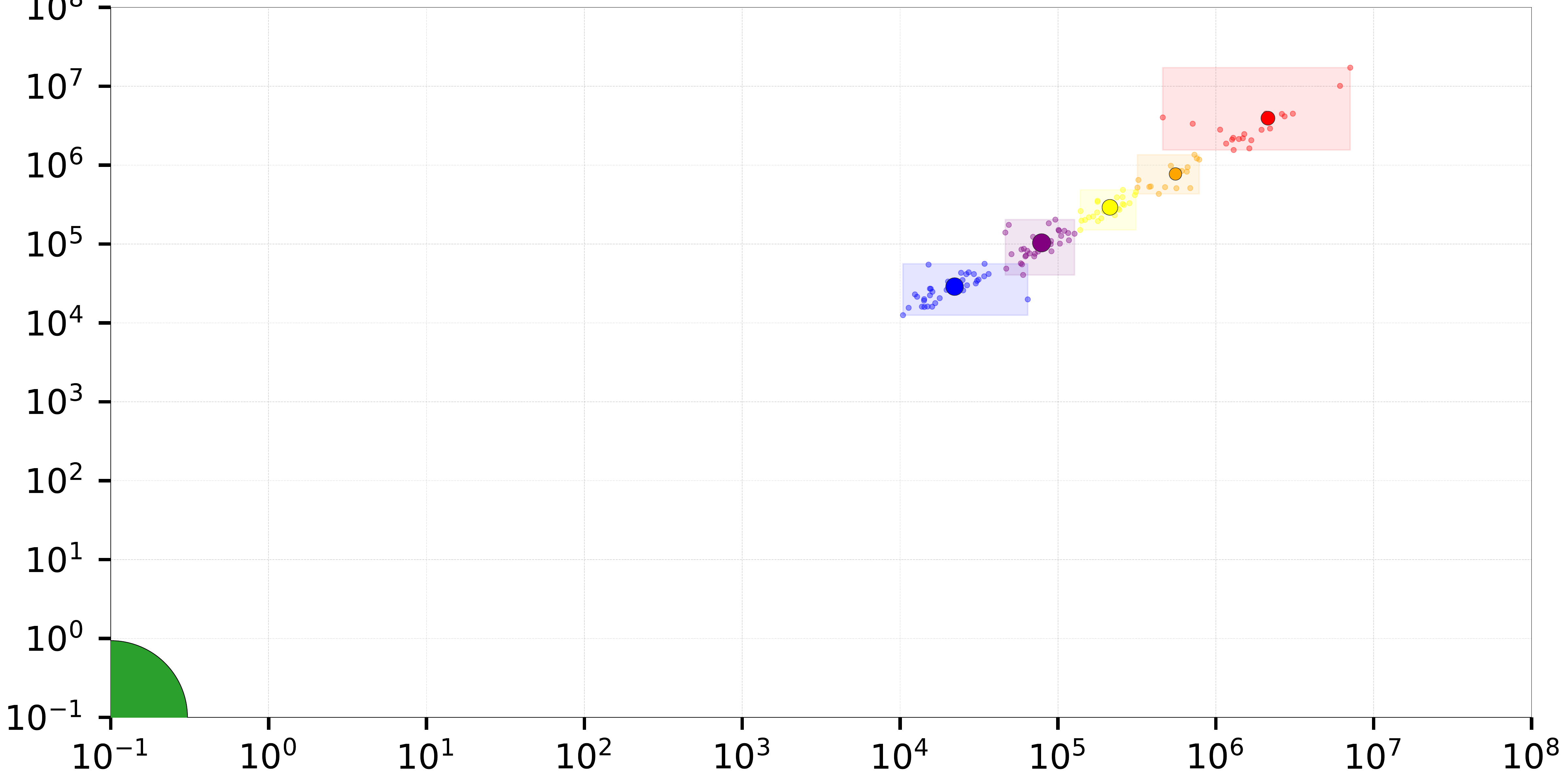}
        \caption{DBpedia-based categories.}
        \label{fig:dbpedia}
    \end{subfigure}
    \qquad
    \begin{subfigure}[b]{0.41\textwidth}
        \centering
        \includegraphics[width=0.80\linewidth]{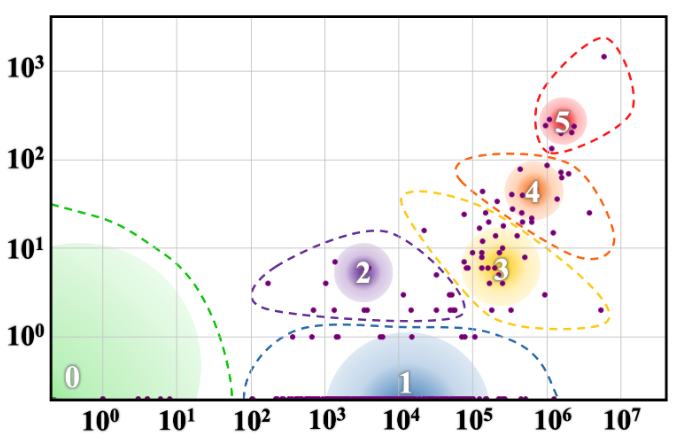}
        \caption{Joshi et al.~\cite{joshi_state_2020} NLP categories.}
        \label{fig:joshicat}
    \end{subfigure}
    \caption[]{Language coverage (log-log) in BabelNet, Wikidata, DBpedia, and in NLP corpora as obtained by Joshi et al.~\cite{joshi_state_2020}. The x-axis represents the number of articles in Wikipedia language editions, while the y-axis represents the count of language-tagged entities in KGs, or the number of available annotated corpora for languages. 
    Clusters and colors from \cite{joshi_state_2020}: \colorcircle{clusterred}{5} Winner, \colorcircle{clusterorange}{4} Underdogs, \colorcircle{clusteryellow}{3} Rising Stars, \colorcircle{clusterpurple}{2} Hopefuls, \colorcircle{clusterblue}{1} Scrapping-Bys, \colorcircle{clustergreen}{0} Left-Behinds.
    }
    \label{fig:clusters_comparison}
\end{figure}

The Natural Language Processing (NLP) community has previously attempted to categorize language digital coverage using qualitative criteria encompassing the quantity of available data or linguistic knowledge~\cite{joshi_state_2020,nigatu_zenos_2024}. 
Joshi et al.~\cite{joshi_state_2020} were the first to propose a quantitative categorization using, for each language, the quantity of annotated corpora available in this language in LDC\footnote{\url{https://catalog.ldc.upenn.edu/}} and ELRA\footnote{\url{https://catalogue.elra.info}}, vs. the quantity of articles available in the language-specific edition of Wikipedia.
%\footnote{\url{https://www.wikipedia.org/}} (Note Pierre: footnote pas utile)
We propose to extend their methodology to obtain a first LOD categorization comparable to theirs. 
Accordingly, for each language, we consider two variables: (i) similarly to~\cite{joshi_state_2020}, the number of articles in its Wikipedia edition, as Wikipedia is often considered a premium source of knowledge to build or enrich KGs; and (ii) the number of entities (synsets for BabelNet) tagged with the target language in a given LOD KG. 
We then apply the k-means algorithm to languages based on these two measures, with $k=6$, and annotated them based on Joshi al.~\cite{joshi_state_2020} language categories description.
The resulting classification reflects both the structured (KG) and unstructured (Wikipedia) language coverage, enabling to devise different uni- or multi-modal enrichment strategies between text and KGs. 
Three multilingual and structurally distinct LOD KGs were primarily used as sources to quantify entity coverage: 
(i) DBpedia,%\footnote{\url{https://www.dbpedia.org}},
(ii) Wikidata, %\footnote{\url{https://www.wikidata.org/wiki/Wikidata:Main_Page}}, 
and (iii)~BabelNet. %\footnote{\url{https://babelnet.org/}}.
This selection was made to capture a representative diversity of the LOD ecosystem in terms of structure and construction paradigm, but is intended to be extended to a broader set of multilingual LOD KGs in future work.
The scope of the study is, as for~\cite{joshi_state_2020}, restricted to written languages of the World Atlas of Language Structures (WALS)\footnote{\url{https://wals.info/languoid}\label{footnote:wals}}. 
The resulting categorization is displayed in Figure~\ref{fig:clusters_comparison}, with detailed figures also available online\footnote{\label{footnote:website}\url{https://nembengue.github.io/language_digital_coverage_lod/}}.

A comparative analysis of the three major LOD KGs (Figure~\ref{fig:clusters_comparison}) reveals heterogeneous distributions across sources. $(i)$ DBpedia exhibits a near-linear, stratified distribution, reflecting its direct dependence on Wikipedia infobox extraction. $(ii)$ Wikidata demonstrates a strongly left-divergent distribution, where languages tend to have substantially more entities than corresponding Wikipedia articles. $(iii)$ BabelNet presents a more compact distribution overall, with categories that also lean toward left-divergency.

Across all LOD KGs and NLP language categorizations, one consistent pattern emerges: the persistent dominance of the \textit{Left-Behinds} category, underscoring the systemic exclusion of the vast majority of the world’s written languages documented in WALS. Beyond this shared observation, however, the application of Joshi et al.'s categorization to LOD remains limited. Although NMI scores (0.63 for DBpedia, 0.60 for BabelNet, and 0.56 for Wikidata) indicate partial overlap, largely driven by the prevalence of \textit{Left-Behinds}, they obscure deeper mismatches, including shifts in language distribution across categories and the collapse of category boundaries when transferred from NLP to LOD.

Furthermore, because Joshi et al.’s categorization is derived from the NLP prism, it fails to capture LOD specificities that suggest different methodological pathways to improve the language coverage. In particular, right-divergencies, where Wikipedia coverage dominates, point to opportunities for automatic knowledge extraction; left-divergencies, where entity coverage exceeds textual resources, call for KG verbalization approaches; and near-linear distributions, as observed in DBpedia, make cross-lingual transfer the most suitable strategy.

Overall, these findings motivate the need for a dedicated categorization framework, specifically tailored to the divergence patterns observed in LOD and flexible enough to reflect the heterogeneity of the language distribution witnessed in LOD.

\section{A Formal Approach for Language Categories in LOD}

To account for the heterogeneity within LOD sources, we systematize the definition of LOD language categories.
%Consider an aggregation of LOD KGs (or one LOD KG) and an aggregation of text corpora (or one corpus such as Wikipedia).
Consider one or more LOD KGs and one or more text corpora (e.g. Wikipedia).
We rely on quantiles of the value distributions across the two introduced dimensions: the number of language-tagged entities in LOD KGs, and the number of language-specific texts in the corpora.

Formally, let $L$ be the set of written languages under WALS, $\llod \subseteq L$ the subset of languages present in the considered LOD KGs, $\ltxt \subseteq L$ the subset of languages present in the considered text corpora, and $L^* =\llod \cap \ltxt$ their intersection.
For each $l \in L^*$, $E(l)$ denotes the set of entities tagged with $l$, and $W(l)$ the set of articles written in $l$, with $|E(l)|$ and $|W(l)|$ their respective cardinalities. 
We then denote $D_E = \big(|E(l_i)|\big)_{l_i \in L^*}$ and $D_W = \big(|W(l_i)|\big)_{l_i \in L^*}$ the \emph{empirical distributions} of entity and article counts, across all languages in $L^*$. By introducing the first and third \emph{empirical quartiles} as $Q_1(D_E)$, $Q_3(D_E)$, $Q_1(D_W)$, and $Q_3(D_W)$, four LOD language categories emerge as follows.

\begin{definitionbox}
\begin{definition}[LOD Language Categories] \\
%\begin{description}
\textbf{Missing Languages.} \label{def:abs} Languages absent from considered KGs or corpora, i.e., $l \in L \text{ and } l \notin L^*$

\textbf{Low-Resource Languages.} Languages in the lower tail of both entity-count and article-count distributions, \\ i.e.,
$l \in L^*
\text{, }
|E(l)| < Q_1(D_E) 
\text{ and }
|W(l)|< Q_1(D_W)
$

\textbf{Medium-Resource Languages.} \label{def:med} Languages in the intermediate range of both entity-count and article-count distributions, \\ i.e.,
$l \in L^*
\text{, }
|E(l)| \in [Q_1(D_E), Q_3(D_E)]
\text{ and }
|W(l)| \in [Q_1(D_W), Q_3(D_W)]$

\textbf{High-Resource Languages.} \label{def:high} Languages in the upper tail of  both entity-count and article-count distributions, \\ i.e.,
$
l \in L^*
\text{, }
|E(l)| > Q_3(D_E) 
\text{ and }
|W(l)| > Q_3(D_W)
$

%\end{description}
\end{definition}
\end{definitionbox}

\begin{figure}[htbp]
    \centering
    \includegraphics[width=0.55\linewidth]{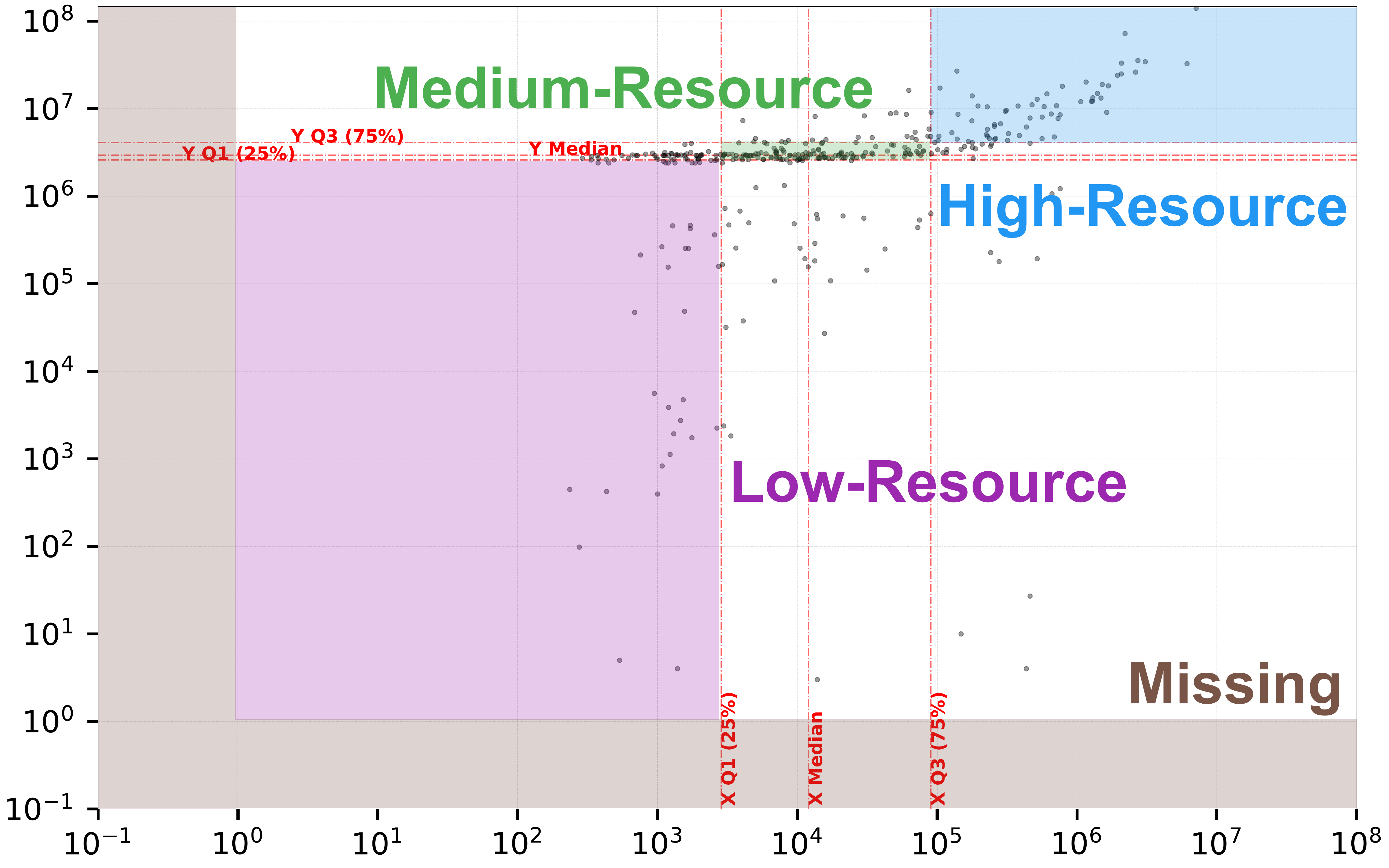}
    \caption{Language coverage (log-log) in the aggregated LOD KGs. 
    The x-axis represents the count of articles in the Wikipedia language editions, and the y-axis the count of language-tagged entities in the KGs. }
    \label{fig:global2} 
\end{figure}

As an illustration, we applied this categorization to the aggregation of DBpedia, Wikidata, and BabelNet as KG LODs, and Wikipedia as a text corpus.
In this preliminary work, we do not perform entity deduplication, and thus Figure~\ref{fig:global2} (also online\footnote{\url{https://nembengue.github.io/language_digital_coverage_lod/}}) presents an optimistic view of language coverage under the assumption that all sources contain distinct entities.
Results reveal that languages are mostly Medium-Resourced, while several languages remain unclassified. 
Future work will refine this categorization with entity deduplication, more granularity, and categories that reflect completion opportunities. 

\section{Conclusion}

The digital divide between high- and low-resource languages, amplified by emerging tools such as Generative AI, is a growing concern.
One solution lies in leveraging multilingual LOD KGs and cross-lingual transfer, which requires understanding language coverage in LOD and identifying target low-resource languages.
To this end, we first replicate a categorization framework established within the NLP community using the number of Wikipedia articles and the number of labeled entities available for each language. We then analyze these categories across three multilingual LOD KGs: BabelNet, DBpedia, and Wikidata. Our findings show that while LOD and NLP approaches align in the ratio of uncovered written languages, other NLP categories are not directly transferable to the LOD context. Thus, we propose a formal framework for building comparable language categories with diverse LOD KGs sources. Our four proposed language categories and definitions are a stepping stone toward addressing digital language coverage through adequate completion strategies.

\paragraph{Acknowledgement.}
This work has been supported by the French government, through the 3IA Côte d’Azur Investments in the project managed by the National Research Agency (ANR) with the reference number ANR-23-IACL-0001, and through the France 2030 investment plan managed by the National Research Agency (ANR), as part of the Initiative of Excellence Université Côte d’Azur under reference number ANR- 15-IDEX-01.
Experiments presented in this paper were carried out using the Grid'5000 testbed, supported by a scientific interest group hosted by Inria and including CNRS, RENATER and several Universities as well as other organizations (see \url{https://www.grid5000.fr}). 
This publication is based upon work from COST Action CA23147 GOBLIN - Global Network on Large-Scale, Cross-domain and Multilingual Open Knowledge Graphs, supported by COST (European Cooperation in Science and Technology, \url{https://www.cost.eu}).

\bibliographystyle{splncs04}
\bibliography{sources}

\end{document}